%% file: main.tex
\documentclass[10pt,twocolumn,letterpaper]{article}

\usepackage{cvpr}              % To produce the CAMERA-READY version
\input{preamble}

\definecolor{cvprblue}{rgb}{0.21,0.49,0.74}
\usepackage[pagebackref,breaklinks,colorlinks,allcolors=cvprblue]{hyperref}

\usepackage{booktabs,multirow,makecell}
\usepackage[table]{xcolor}
\usepackage{pifont} % Add this in your preamble

\newcommand{\cmark}{\ding{51}}%
\newcommand{\xmark}{\ding{55}}%

\definecolor{splattalkblue}{RGB}{209,244,247} % light blue for SplatTalk rows
\newcommand{\NA}{--}

\def\confName{CVPR}
\def\confYear{2026}
\usepackage{pifont} 
\title{POMA-3D: The Point Map Way to 3D Scene Understanding}

\author{Ye Mao \quad Weixun Luo \quad Ranran Huang \quad Junpeng Jing\footnotemark[2] \quad Krystian Mikolajczyk \\[8pt]
Imperial College London\\
}

\begin{document}

\twocolumn[{
\renewcommand\twocolumn[1][]{#1}
\maketitle
\begin{center}
    \centering
    \vspace{-0.9em}\includegraphics[width=1\linewidth,page=1]{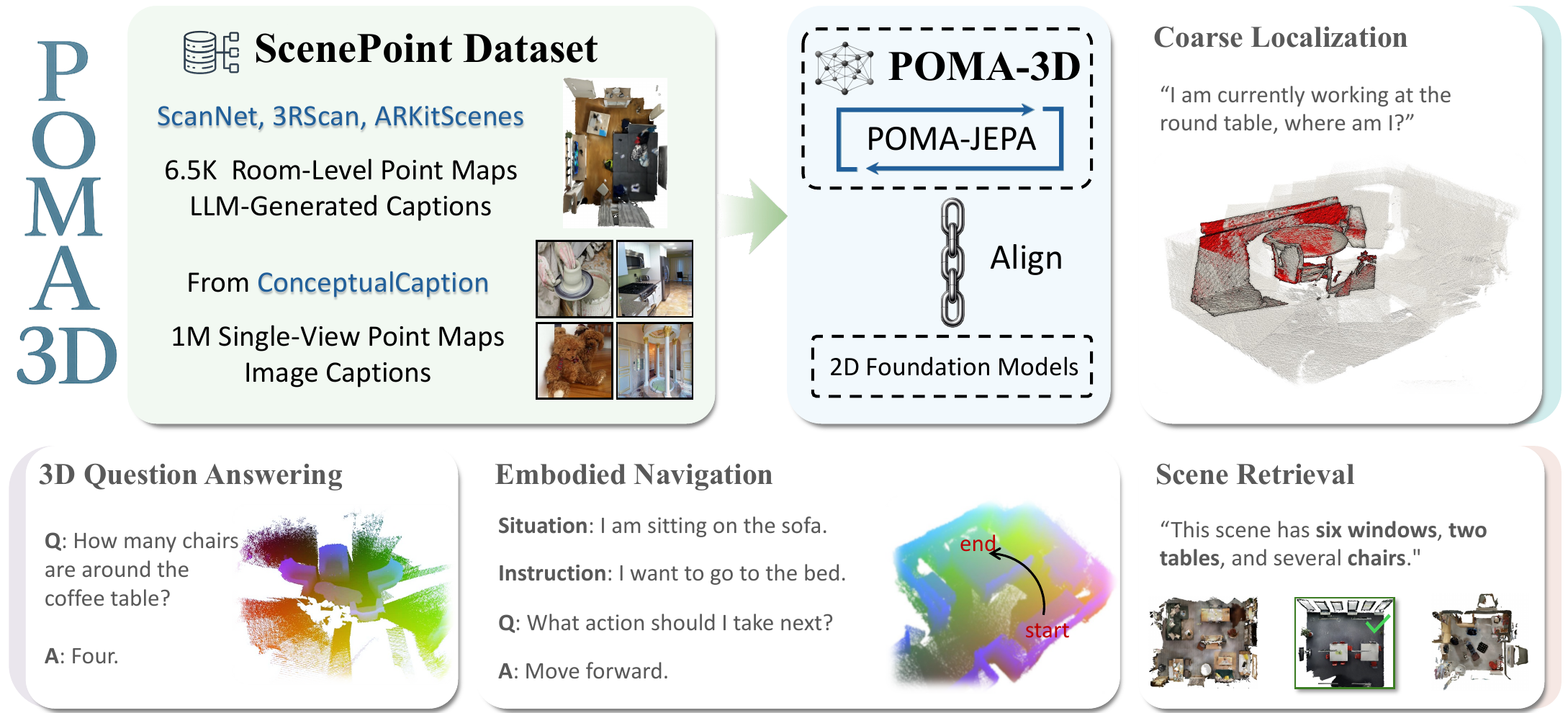}
    % \vspace{-1.2em}
    \captionsetup{hypcap=false} % Disable hypcap for this specific caption
    \captionof{figure}{\textbf{Overview of POMA-3D.} POMA-3D is a self-supervised 3D model pretrained on the large-scale point map dataset ScenePoint via alignment with 2D foundation models and the POMA-JEPA objective. The 3D features from pretrained POMA-3D generalize to diverse 3D scene understanding tasks, including 3D question answering, embodied navigation, scene retrieval, and coarse localization.}
    \label{fig:1}
\end{center}
}]
\renewcommand{\thefootnote}{\fnsymbol{footnote}}
\footnotetext[2]{Corresponding author: \texttt{j.jing23@imperial.ac.uk}}
\input{sec/0_abstract}    
\input{sec/1_intro}

\input{sec/2_related}

\input{sec/3_scenepoint}
\input{sec/4_poma3d}
\input{sec/5_result}
\input{sec/6_conclusion}

\section{Acknowledgments} Supported by the Imperial President’s Scholarship.
\clearpage

{
    \small
    \bibliographystyle{ieeenat_fullname}
    \bibliography{main}
}

%     \small
%     \bibliographystyle{ieeenat_fullname}
%     \bibliography{appbib}
% }

\end{document}

%% file: preamble.tex
\usepackage{mathtools}

%% file: sec/0_abstract.tex
\begin{abstract}
In this paper, we introduce POMA-3D, the first self-supervised 3D representation model learned from point maps. Point maps encode explicit 3D coordinates on a structured 2D grid, preserving global 3D geometry while remaining compatible with the input format of 2D foundation models. To transfer rich 2D priors into POMA-3D, a view-to-scene alignment strategy is designed. Moreover, as point maps are view-dependent with respect to a canonical space, we introduce POMA-JEPA, a joint embedding-predictive architecture that enforces geometrically consistent point map features across multiple views. Additionally, we introduce ScenePoint, a point map dataset constructed from 6.5K room-level RGB-D scenes and 1M 2D image scenes to facilitate large-scale POMA-3D pretraining.
Experiments show that POMA-3D serves as a strong backbone for both specialist and generalist 3D understanding. It benefits diverse tasks, including 3D question answering, embodied navigation, scene retrieval, and coarse localization, all achieved using only geometric inputs (i.e., 3D coordinates).
Overall, our POMA-3D explores a \textbf{point map way to 3D scene understanding}, addressing the scarcity of pretrained priors and limited data in 3D representation learning. Code and dataset are publicly available at \href{https://matchlab-imperial.github.io/poma3d/}{https://matchlab-imperial.github.io/poma3d/}.
\end{abstract}

%% file: sec/1_intro.tex
\section{Introduction}
\label{sec:intro}
Understanding 3D scenes is fundamental for perceiving and interacting with the physical world, forming the basis of contextual intelligence in AR systems and embodied agents~\cite{li2025embodied,duan2022survey}. Early 3D understanding models were specialist, targeting specific 3D tasks such as instance segmentation~\cite{jiang2020pointgroup,hou20193d}, visual grounding~\cite{chen2020scanrefer,achlioptas2020referit3d}, or visual question answering (QA)~\cite{azuma2022scanqa,ma2022sqa3d}. Recently, generalist 3D models~\cite{zheng2025video,zhu2024llava,fan2025vlm,fu2024scene,chen2024ll3da,wang2025ross3d} have emerged, leveraging large language models (LLMs) to achieve holistic 3D understanding across diverse tasks within a unified framework.

Both specialist and generalist models hinge on robust spatial representations. 3D vision-language learning (VLL) provide them through contrastive objectives without relying on downstream annotations. Early 3D VLL methods align trainable 3D encoders with frozen 2D vision-language models (e.g., CLIP~\cite{radford2021learning}). This cross-modal alignment enables 3D encoders to inherit rich knowledge from 2D counterparts, enabling zero-shot tasks such as object classification, retrieval, and detection~\cite{liu2023openshape, xue2023ulip, mao2024opendlign, zhou2023uni3d, xue2024ulip, qi2024shapellm}. Subsequent methods~\cite{zhu20233d, jia2024sceneverse} extend this paradigm from object to scene-level representation learning. However, existing 3D VLL methods have yet to achieve a breakthrough comparable to the CLIP moment in 2D understanding.

The key reason is that these methods primarily utilize point clouds~\cite{jia2024sceneverse,zhu20233d}, depth maps~\cite{mao2024opendlign,huang2023clip2point}, or 3D Gaussian splatting~\cite{li2025scenesplat,thai2025splattalk} for alignment, all of which differ substantially from pretrained 2D representations. \textit{In this paper, we argue that point maps can provide a superior intermediate 2D–3D modality for better alignment.} This is enabled by recent advances in feed-forward 3D reconstruction models~\cite{wang2024dust3r,wang2025vggt,leroy2024grounding,huang2025no}.
Unlike other 3D inputs, point maps encodes pixel-to-3D correspondences in a 2D grid, naturally naturally matching the data format of 2D foundational models. Moreover, multi-view point maps are defined in canonical world coordinates, preserving the same global geometry as point clouds. These properties collectively make point maps a promising modality that retains rich 3D information while aligning closer with 2D knowledge. 

Motivated by these properties, we introduce POMA-3D, the first self-supervised 3D learning framework built upon point maps, as illustrated in Fig.~\ref{fig:1}. To enable large-scale pretraining, we construct ScenePoint, a point map dataset comprising over 6.5K room-level indoor scenes paired with LLM-generated descriptions. In addition, we convert 1M images from image-caption datasets into single-view point maps using the VGGT model~\cite{wang2025vggt} for 3D learning. Building on ScenePoint, POMA-3D is pretrained with a view-to-scene vision–language alignment objective, encouraging the model to learn CLIP-aligned point map embeddings. To ensure feature consistency across viewpoints, we further design the Point Map Joint Embedding-Predictive Architecture (POMA-JEPA) as an additional training objective. Unlike traditional JEPAs~\cite{assran2023self, bardes2023v,assran2025v}, POMA-JEPA explicitly handles the order-agnostic nature of point maps in the world coordinate frame by enforcing permutation-invariant embedding prediction. The overall training follows a two-stage paradigm: a warm-up stage using 2D image scenes, followed by a main stage using indoor room scenes.

We evaluate the generalizability of POMA-3D across diverse 3D scene understanding tasks, including 3D question answering, embodied navigation, and scene retrieval. After pretraining, POMA-3D gains the ability to accurately locate the agent’s active region from textual queries in a zero-shot setting, a task we term coarse localization (see Fig.~\ref{fig:1}). When used as a backbone for both specialist and generalist 3D models, POMA-3D consistently outperforms existing state-of-the-art 3D VLL methods, even without color information—using only pure 3D coordinates. Notably, our results demonstrate that leveraging 2D vision–language data significantly benefits POMA-3D pretraining, highlighting a promising direction toward addressing the long-standing data scarcity challenge in building foundation models for 3D understanding.

Our contributions can be summarized as follows:
\begin{itemize}
    \item We propose POMA-3D, the first self-supervised model that learns 3D scene representations from point maps.
    \item We present ScenePoint, a large-scale point map dataset comprising 6.5K room-level and 1M image scenes for POMA-3D two-stage pretraining. 
    \item We design a view-to-scene vision–language alignment and  POMA-JEPA as training objectives to learn CLIP-aligned and multi-view consistent point map features.
\end{itemize}

%% file: sec/2_related.tex
\section{Related Work}
\label{sec:related}
\noindent \textbf{From Specialist to Generalist 3D Models.}     
Existing 3D scene understanding methods fall into specialist and generalist paradigms. Specialist models~\cite{jiang2020pointgroup,hou20193d,chen2020scanrefer,qi2017pointnet,guo2023viewrefer,zhao2021point,huang2022multi} are tailored for individual tasks such as segmentation or grounding, achieving strong performance but limited cross-task generalization. Building on advances in large language models (LLMs), recent 3D generalist models aim to unify perception and reasoning across modalities. Early efforts such as 3D-LLM~\cite{hong20233d} adapt LLMs to process 3D features from rendered images, while Chat3D~\cite{wang2023chat} and LEO~\cite{huang2023embodied} enhance 3D reasoning by integrating object-centric representations from off-the-shelf 3D detectors. LLaVA-3D~\cite{zhu2024llava} extends 2D visual instruction tuning to 3D via voxelized patch aggregation, and Video-3D LLM~\cite{zheng2025video} incorporates 3D cues into video-based representations. SR-3D~\cite{cheng20253d} is the most related work to ours, using point maps to build a generalist 3D model. However, it uses point maps only as a source of 3D positional encoding rather than pretraining a dedicated encoder to learn point map representations. In this work, we propose POMA-3D, which learns generalizable point map features that effectively benefit both specialist and generalist models across diverse 3D understanding tasks. 

\noindent \textbf{3D Vision-Language Learning.} 
Existing 3D vision-language learning (VLL) methods align 3D data (e.g., point clouds, depth maps, voxel grids), multi-view images, and text through contrastive objectives, transferring rich semantic priors from CLIP into 3D domains. Early works such as ULIP~\cite{xue2023ulip}, OpenShape~\cite{liu2023openshape}, OpenDlign~\cite{mao2024opendlign}, and Uni3D~\cite{zhou2023uni3d} demonstrate that CLIP-based supervision produces strong 3D object representations for open-world recognition and retrieval. However, these approaches remain object-centric, limiting their ability to capture holistic scene semantics. To address this limitation, 3D-VisTA~\cite{zhu20233d} extends VLL to 3D scenes by aligning scene-level point cloud features with scene captions, while SceneVerse~\cite{jia2024sceneverse} scales this approach to a large corpus of 3D scenes, yielding promising results in 3D visual grounding. More recently, SceneSplat~\cite{li2025scenesplat} explores VLL on 3D Gaussian splats, producing continuous 3D features that enhance scene segmentation. Building on this line of research, POMA-3D also adopts vision-language alignment as a key training objective. Unlike prior scene-based methods that focus on object- or scene-level alignment, POMA-3D aligns point map view and scene representations, as summarized in Tab.~\ref{tab:1}.

\noindent \textbf{3D Scene Pretraining Datasets.}
Collecting large-scale 3D scene data for 3D vision-language learning remains a major challenge due to the high cost and prolonged scanning time required by 3D sensing devices. Widely used datasets such as ScanNet~\cite{dai2017scannet}, 3RScan~\cite{wald2019rio}, ARKitScenes~\cite{baruch2021arkitscenes}, HM3D~\cite{ramakrishnan2021habitat}, and MultiScan~\cite{mao2022multiscan} contain only thousands of scenes, which is much smaller than the billion-scale image text corpora used for 2D pretraining. To mitigate this limitation, several works~\cite{jia2024sceneverse} have leveraged synthetic 3D scene datasets such as Structured3D~\cite{zheng2020structured3d} and ProcTHOR~\cite{deitke2022️}. However, the limited realism of synthetic scenes restricts their effectiveness in learning generalizable real-world representations. In this work, we leverage 2D vision–language datasets for 3D learning by using a feed-forward 3D model to convert images into pseudo point maps, enabling scalable 3D vision-language learning.

%% file: sec/3_scenepoint.tex
\section{ScenePoint Dataset}

% in preamble: \usepackage{pifont}

\begin{table}[t]\addtolength{\tabcolsep}{-5.5pt}
\centering
\small
\caption{Comparison of ScenePoint with existing indoor 3D vision-language datasets. “CC" for ConceptualCaptions~\cite{sharma2018conceptual}.}
\label{tab:dataset_comparison}
\vspace{-1em}
\begin{tabular}{l|ccc}
\toprule
Dataset/Attribute & SceneScribe~\cite{zhu20233d} & SceneVerse~\cite{jia2024sceneverse} & ScenePoint \\
\midrule
ScanNet~\cite{dai2017scannet}         & \cmark & \cmark & \cmark \\
ARKitScenes~\cite{baruch2021arkitscenes}    & \xmark & \cmark & \cmark \\
HM3D~\cite{ramakrishnan2021habitat}      & \xmark & \cmark & \xmark \\
3RScan~\cite{wald2019rio}        & \cmark & \cmark & \cmark \\
MultiScan~\cite{mao2022multiscan}    & \xmark & \cmark & \xmark \\
Structured3D~\cite{zheng2020structured3d} & \xmark & \cmark & \xmark \\
ProcTHOR~\cite{deitke2022️}    & \xmark & \cmark & \xmark \\
CC~\cite{sharma2018conceptual} & \xmark & \xmark & \cmark \\
\midrule
Caption Detail & Object, Scene & Object, Scene & \textbf{View, Scene} \\
Room-level Scene & 3.0K & 68K & \textbf{6.5K} \\
Single-view Scene & – & – & \textbf{1M} \\
\bottomrule
\end{tabular}
\label{tab:1}
\vspace{-0.5em}
\end{table}

We introduce ScenePoint, a dataset of aligned triplets consisting of point maps, images, and captions of 3D indoor scenes for vision-language point map pretraining. As summarized in Tab.~\ref{tab:1}, ScenePoint integrates diverse RGB-D scene datasets, each annotated with both LLM-generated view-level and scene-level captions. In addition, it includes single-view scenes generated from image caption datasets. 

\subsection{Point Map Curation}
Multi-view point maps are constructed from RGB-D room videos by sampling 32 frames per video. Following the maximum sampling coverage strategy of Video-3D-LLM~\cite{zheng2025video}, the selected frames capture the maximum spatial extent of each scene. Each point map view $P_i$ is generated from its corresponding depth map $D_i \in \mathbb{R}^{H \times W}$ using the intrinsic matrix $K$ and the extrinsic parameters $E_i = [R_i \,|\, t_i]$, where $R_i$ and $t_i$ represent rotation and translation, respectively. The resulting point map $P_i \in \mathbb{R}^{H \times W \times 3}$ preserves the spatial resolution of $D_i$, where each pixel in $(u, v)$ stores its 3D coordinate $(x, y, z)$ as:
\begin{equation}
\begin{bmatrix}
x \\ y \\ z
\end{bmatrix}
=
R_i \left( D_i(u,v) K^{-1}
\begin{bmatrix}
u \\ v \\ 1
\end{bmatrix}
\right) + t_i.
\label{eq:1}
\end{equation}

Single-view point maps are generated from images in the ConceptualCaptions dataset~\cite{sharma2018conceptual} using Eq.~\ref{eq:1}, where depth maps and camera parameters are predicted by the depth and pose heads of the VGGT model~\cite{wang2025vggt}.

\subsection{Language Generation}
Language annotations are provided at both the view and scene levels of the point maps. For room-level point maps, each view is paired with captions generated by InternVL3-14B~\cite{zhu2025internvl3}, using the corresponding RGB image as input. Specifically, 15 candidate captions $\{c_j\}_{j=1}^{15}$ are first produced, and their cosine similarities with the corresponding FG-CLIP~\cite{xie2025fg} image embedding $f^{\text{img}}$ are computed as
$s_j = \cos(f^{\text{img}}, f^{\text{text}}_{c_j})$.
The top-5 captions with the highest $s_j$ are retained as the final view-level annotations. Scene-level captions for room-level point maps are adopted from SceneVerse~\cite{jia2024sceneverse}. For single-view point maps, the original image captions are directly used as their view-level annotations.

\subsection{Dataset Statistics}
Tab.~\ref{tab:1} shows that ScenePoint comprises 6,562 indoor scenes collected from three RGB-D datasets, including 1,499 from ScanNet~\cite{dai2017scannet}, 1,204 from 3RScan~\cite{wald2019rio}, and 3,850 from ARKitScenes~\cite{baruch2021arkitscenes}. Unlike previous datasets, ScenePoint does not include any synthetic 3D scenes. Instead, it incorporates 1M image scenes sampled from the ConceptualCaptions~\cite{sharma2018conceptual} dataset. 

%% file: sec/4_poma3d.tex
\begin{figure*}[!ht]
    \centering
    \includegraphics[width=1.0\linewidth]{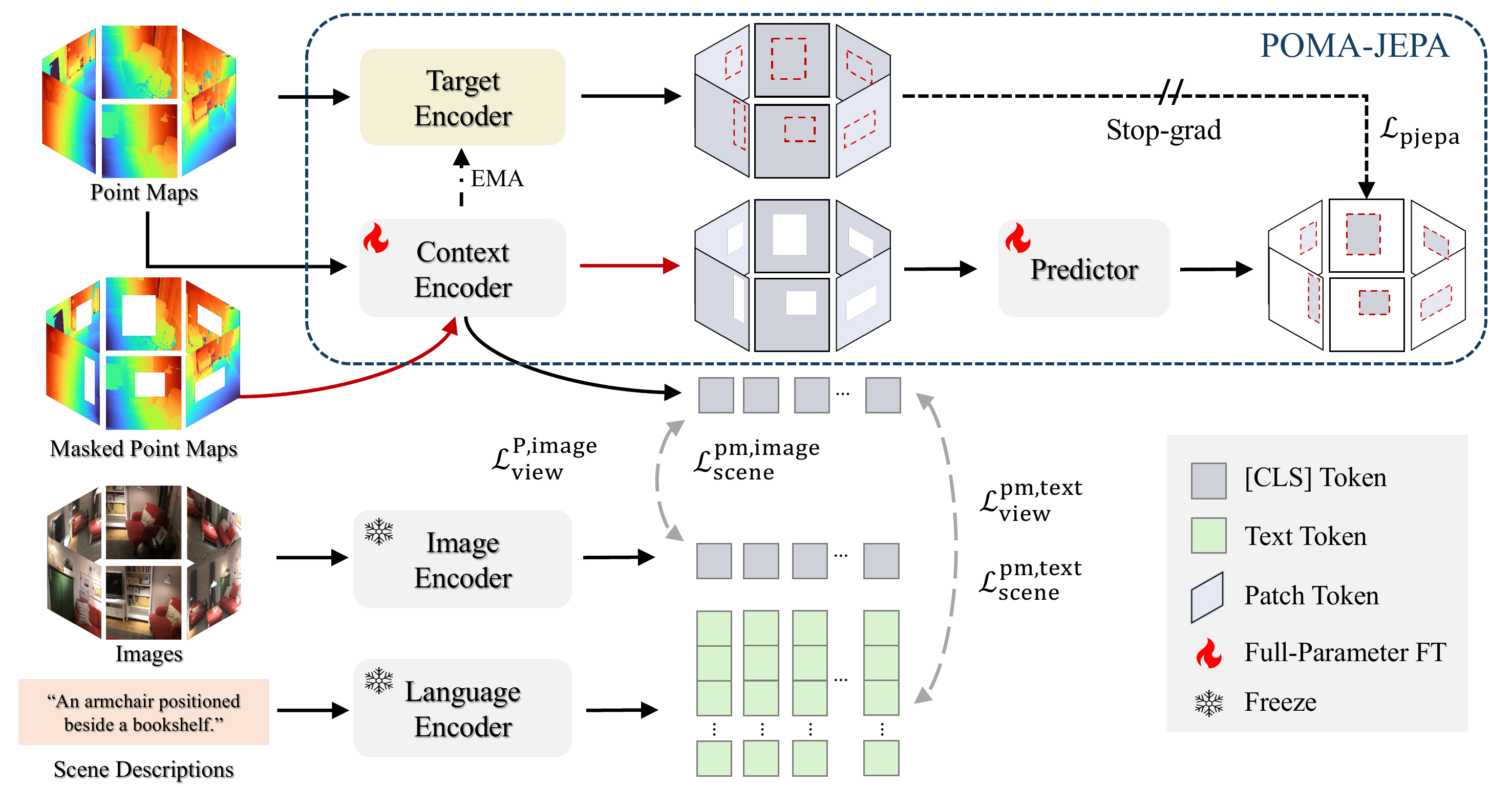}
    \vspace{-2.1em}
    \caption{\textbf{Overview of the POMA-3D pretraining.} 
POMA-3D is pretrained with two objectives: (1) aligning \texttt{[CLS]} embeddings from the point map context encoder with image and text embeddings from the frozen FG-CLIP using $\mathcal{L}_{\text{view}}$ and $\mathcal{L}_{\text{scene}}$, and (2) reconstructing masked point map embeddings from the target encoder using unmasked embeddings from the context encoder via a predictor optimized by $\mathcal{L}_{\text{pjepa}}$. The target encoder is updated via EMA of the context encoder, and its final weights are used for downstream 3D understanding.}
    \label{fig:2}
\vspace{-1.0em}
\end{figure*}
\section{POMA-3D}
The overall pretraining framework of POMA-3D is illustrated in Fig.~\ref{fig:2}. Built upon the ScenePoint dataset, POMA-3D is trained with two objectives: (1) \textit{view-to-scene vision–language alignment} (Sec.~\ref{4.1}) to align point map, image, and text embeddings, and (2) \textit{POMA-JEPA} (Sec.~\ref{4.2}) to enforce geometric consistency across multi-view point map embeddings. Frozen image and language encoders are aligned with a trainable point map context encoder. The context encoder, together with a predictor and an EMA-updated target encoder, optimizes POMA-JEPA. The features from the pretrained target encoder will finally be used for downstream analysis.
\subsection {View-to-Scene Vision-Language Alignment} 
\label{4.1}
A scene consists of point maps $P=\{P_i\}_{i=1}^{N_v}$, images $I=\{I_i\}_{i=1}^{N_v}$, view-level captions $V=\{V_i\}_{i=1}^{N_v}$, and optionally a scene-level caption $S$, where $N_v$ denotes the number of views. FG-CLIP~\cite{xie2025fg} is chosen as the backbone for view-to-scene vision–language alignment due to its extended text token capacity, enabling effective modeling of long scene descriptions. Specifically, the alignment incorporates pretrained FG-CLIP image and language encoders $E_I$ and $E_L$. A context point map encoder $E_C$, initialized from $E_I$ and finetuned, is used to bridge 3D and 2D modalities. 

For $i_{th}$ view, the encoders produce modality-specific embeddings: $z_P^{i} = E_C(P_i)$, $z_I^{i} = E_I(I_i)$, and $z_V^{i} = E_L(V_i)$, where $z_P^{i}$, $z_I^{i}$, and $z_V^{i}$ denote the \texttt{[CLS]} token embeddings of the point map, image, and per-view caption, respectively. The view-level alignment encourages each point map embedding to align with its paired image and caption embeddings while contrasting against different views in the batch, selected by the maximum coverage sampling. For any two modalities $M_1$ and $M_2$, the loss is:
\begin{align}
\mathcal{L}_{\text{view}}^{M_1, M_2} = -\frac{1}{2}
\sum_{(i,j)}
\Bigg(
\log\frac{\exp\big({z}_{M_1}^{i}\!\cdot\!{z}_{M2}^{j}/\tau\big)}
{\sum_{k}\exp\big({z}_{M_1}^{i}\!\cdot\!{z}_{M_2}^{k}/\tau\big)}
\nonumber \\[-3.5pt]
{
+\log\frac{\exp\big({z}_{M_1}^{i}\!\cdot\!z_{M_2}^{j}/\tau\big)}
{\sum_{k}\exp\big({z}_{M_1}^{k}\!\cdot\!z_{M_2}^{j}/\tau\big)}
}
\Bigg),
\end{align}
where $(i,j)$ denotes a positive pair, while $(i,k)$ and $(k,j)$ represent negative pairs within the batch, and $\tau$ is the temperature parameter. Specifically, $\mathcal{L}_{\text{view}}^{{P,I}}$ aligns point map and image modalities ($M_1{=}P$, $M_2{=}I$), whereas $\mathcal{L}_{\text{view}}^{{P,V}}$ aligns point map and text modalities ($M_1{=}P$, $M_2{=}V$). The total view loss $\mathcal{L}_{\text{view}} = \mathcal{L}_{\text{view}}^{P,I} + \mathcal{L}_{\text{view}}^{P,V}$.

For scene-level alignment, point map and image embeddings $\{z_P^{i}\}_{i=1}^{N_v}$ and $\{z_I^{i}\}_{i=1}^{N_v}$ within each scene are mean-pooled into scene embeddings $\bar{z}_P$ and $\bar{z}_I$. The scene caption $S$ is encoded as $\bar{z}_S = E_L(S)$.  Each scene’s point map embedding is aligned with its paired image and caption embeddings while contrasting against other scenes in the batch:
\begin{align}
\mathcal{L}_{\text{scene}}^{{M_1,M_2}} = -\frac{1}{2}
\sum_{(i,j)}
\Bigg(
\log\frac{\exp\big(\bar{z}_{M_1}^{i}\!\cdot\!\bar{z}_{M_2}^{j}/\tau\big)}
{\sum_{k}\exp\big(\bar{z}_{M_1}^{i}\!\cdot\!\bar{z}_{M_2}^{k}/\tau\big)}
\nonumber \\[-3.5pt]
+\log\frac{\exp\big(\bar{z}_{M_1}^{i}\!\cdot\!\bar{z}_{M_2}^{j}/\tau\big)}
{\sum_{k}\exp\big(\bar{z}_{M_1}^{k}\!\cdot\!\bar{z}_{M_2}^{j}/\tau\big)}
\Bigg),
\end{align}
\noindent where $\mathcal{L}_{\text{scene}}^{P,I}$ and $\mathcal{L}_{\text{scene}}^{P,S}$ denote scene-level point map–image and point map–caption alignments. $\mathcal{L}_{\text{scene}} = \mathcal{L}_{\text{scene}}^{P,I} + \mathcal{L}_{\text{scene}}^{P,S}.
$
\subsection{POMA-JEPA} \label{4.2}
As shown in Fig.~\ref{fig:2}, the POMA-JEPA pretraining module consists of a context encoder $E_C$, a target encoder $E_T$, and a predictor $f_\theta$. Given multi-view point maps $\{P_i\}_{i=1}^{N_v}$, a random masking function $\mathcal{M}(\cdot)$ is applied to each view to mask a subset of patches. The union of all masked patches across views is denoted as $\Omega_M$, and its visible complement is $\Omega_V$ (see appendix for examples). The context encoder encodes the visible regions from all views to obtain latent features $Z_C = \{E_C(P_i^{\Omega_V})\}_{i=1}^{N_v}$. 
The target encoder processes the complete point maps to produce full patch embeddings $Z_T = \{E_T(P_i)\}_{i=1}^{N_v}$. The predictor takes the concatenated context embeddings from all views to reconstruct the target embeddings of the masked regions, 
$\hat{Z}_T = f_\theta(Z_C)$.  During POMA-JEPA training, the context encoder  $E_C$ continues fine-tuning, while the target encoder $E_T$ is updated as the EMA of the context encoder $E_C$ after each iteration.

Since the merged point maps in the world coordinate frame form a dense point cloud, it naturally inherits the order-agnostic nature of point sets. Consequently, the predicted patch embeddings $\hat{Z}_T$ do not necessarily follow the same spatial order as the target embeddings $Z_T$. The standard JEPA utilizes MSE loss that enforces a one-to-one mapping in a fixed 2D grid, which we find leads to mode collapse in the 3D setting. Here, we define the POMA-JEPA loss $\mathcal{L}_{\text{pjepa}}$ using the Chamfer Distance~\cite{fan2017point}, a widely used metric in masked point cloud modeling~\cite{pang2023masked} for its robustness to minor order misalignments, defined as:
\begin{equation}
\mathcal{L}_{\text{pjepa}}
=\!\!\sum_i\min_j\|\hat{Z}_T^i - Z_T^j\|_2^2
+\!\!\sum_j\min_i\|Z_T^j - \hat{Z}_T^i\|_2^2,
\label{eq:jepa_chamfer}
\end{equation}
where $i,j \in \Omega_M$ denote indices of masked patches.

\subsection{Two-Stage Pretraining} \label{sec:two_stage}
POMA-3D is pretrained in two stages. 
The first \textbf{warm-up stage} performs vision–language alignment on all image-derived single-view point maps, with the total loss defined as $\mathcal{L}_{\text{total}} = \mathcal{L}_{\text{view}}$. 
The \textbf{main stage} jointly optimizes alignment and POMA-JEPA pretraining on multi-view point maps from room-level scenes. 
Both the context and target point map encoders are initialized from the context encoder weights obtained after the warm-up stage. 
The total loss for this stage is defined as:
\begin{equation}
    \mathcal{L}_{\text{total}} = \mathcal{L}_{\text{view}} + \mathcal{L}_{\text{scene}} + \mathcal{L}_{\text{pjepa}}.
\end{equation}

%% file: sec/5_result.tex
% Preamble (add to your preamble if not already present)

% Tab.~0: Pretraining data for different vision-language methods.
% Tab.~1: Main table: ScanQA, SQA3D, Hypo3D, MSQN, Scan2Cap
% Tab.~2: Main table: ScanQA, SQA3D, Hypo3D Uncolored Questions.
% Tab.~3: Ablation Study: warmup, contrastive learning, masking loss.
% Tab.~4: Ablation Study: RGB, RGB + Contrastive, Depth Map, Depth Map + Contrastive, Point Map
% Tab.~5: Ablation Study: Different Number of Views

% TODO List
% LLaVA-OV 7B ScanQA SQA3D Hypo3D MSNN
% Qwen2-VL 7B ScanQA SQA3D Hypo3D MSNN
% SceneVerse Hypo3D MSNN
% 3D-Vista Hypo3D MSNN
% Scan2Cap POMA-3D
% POMA-3D + LLaVA-OV
% POMA-3D > baseline on MSNN and Hypo3D

\definecolor{bestcolor}{RGB}{220,242,220}      % soft mint green – best
\definecolor{secondcolor}{RGB}{236,220,250}    % gentle lavender – second best
\definecolor{lightgrayrow}{RGB}{245,245,245}

\begin{table*}[ht]\addtolength{\tabcolsep}{2.0pt}
\centering
\small

\caption{3D QA results on ScanQA~\cite{azuma2022scanqa}, SQA3D~\cite{ma2022sqa3d}, and Hypo3D~\cite{mao2025hypo3d}, and embodied navigation results on MSNN~\cite{linghu2024multi}.
4 dir./8 dir.:four/eight-directional navigation; C-PC: colored point cloud; RGB: image; RGB-D: image + depth; GS: Gaussian Splat; PM: point map. ‘$\dagger$’ indicates models with $H \times W \times 3$ grid inputs and LoRA-tuned for one epoch from the 2D LLM.
 ‘–’ indicates the metric is inapplicable or the result is unavailable. \colorbox{bestcolor}{\textbf{Best}} and \colorbox{secondcolor}{second-best} 2D LLM-based and specialist model results are highlighted.}

\vspace{-1em}
\begin{tabular}{l c cc cc cc cc}
\toprule
Method & {Modality} 
& \multicolumn{2}{c}{ScanQA} 
& \multicolumn{2}{c}{SQA3D}
& \multicolumn{2}{c}{Hypo3D}
& \multicolumn{2}{c}{MSNN} \\
\cmidrule(lr){3-4} \cmidrule(lr){5-6} \cmidrule(lr){7-8} \cmidrule(lr){9-10}
 &  & EM@1 & EM@10 & EM@1 & EM@10 & EM@1 & EM@10 & 4 dir. & 8 dir. \\
\midrule
\rowcolor{lightgrayrow}
\multicolumn{10}{l}{\emph{{3D LLM Models}}}\\
LEO~\cite{huang2023embodied}          & C-PC     & 24.5 & --   & 50.0 & --   & 16.2  & --  & -- & --  \\
LLaVA-3D~\cite{zhu2024llava} & RGB-D  & 27.0 & --   & 55.6 & --   & 33.1& --  & 22.9 & 12.3 \\
Video-3D 
LLM~\cite{zheng2025video} & RGB-D  & 30.1 & --   & 58.6 & --   & \NA  & \NA & -- & -- \\
\midrule
\rowcolor{lightgrayrow}
\multicolumn{10}{l}{\emph{{2D LLM-based Models (Pretrained/LoRA-tuned)}}}\\
Qwen2.5-VL-7B~\cite{bai2025qwen2} & RGB & \cellcolor{bestcolor}\textbf{23.7} & --   & \cellcolor{secondcolor}47.8 & --   & 30.9 & --  & 21.8 & 2.87 \\
LLaVA-OV-7B~\cite{li2024llava}   & RGB  & 20.8 & --   & 47.7 & --   & \cellcolor{secondcolor}33.2 & --  & \cellcolor{secondcolor}24.0 & \cellcolor{secondcolor}5.83 \\
SplatTalk$^{\dagger}$~\cite{thai2025splattalk}   & RGB(GS) & \cellcolor{secondcolor}22.4 & --   & 47.6 & --   & \NA  & \NA & \NA & \NA \\
\textbf{POMA-3D}$_{\text{llm}}^{\dagger}$ & PM & 21.3 & -- &
\cellcolor{bestcolor}\textbf{51.6} & -- &
\cellcolor{bestcolor}\textbf{35.9} & \NA &
\cellcolor{bestcolor}\textbf{36.9} & 
\cellcolor{bestcolor}\textbf{21.4} \\
\midrule 
\rowcolor{lightgrayrow}
\multicolumn{10}{l}{\emph{Specialist Models}}\\
ScanQA~\cite{azuma2022scanqa}             & C-PC  & 21.1 & --   & 47.2 & --   & \NA  & \NA & \NA & \NA \\
% ScanRefer+MCAN~\cite{chen2020scanrefer}  & C-PC  & 18.6 & --   & \NA  & \NA  & \NA  & \NA & \NA & \NA \\
SQA3D~\cite{ma2022sqa3d}               & C-PC  & \NA  & \NA  & 46.6 & --   & \NA  & \NA & \NA & \NA \\
3D-ViSTA~\cite{zhu20233d}          & C-PC  & \cellcolor{secondcolor}22.4 & \cellcolor{secondcolor}52.1 & 48.5 & 85.6 & 31.0 & 81.2 & \cellcolor{secondcolor}39.9 & 20.1 \\
SceneVerse~\cite{jia2024sceneverse}     & C-PC  & \cellcolor{bestcolor}\textbf{22.7} & 51.5 & \cellcolor{secondcolor}49.9 & 85.0 & \cellcolor{secondcolor}31.6 & 80.3 & 36.0 & 19.5 \\
FG-CLIP~\cite{xie2025fg}            & PM  & 20.9 & 49.9 & 49.5 & \cellcolor{secondcolor}89.7 & 31.1 & \cellcolor{secondcolor}82.1 & 39.3 & \cellcolor{secondcolor}20.4 \\
\textbf{POMA-3D$_{\text{spec}}$}              & \textbf{PM}  &
22.3 &
\cellcolor{bestcolor}\textbf{52.3} &
\cellcolor{bestcolor}\textbf{51.1} &
\cellcolor{bestcolor}\textbf{91.2} &
\cellcolor{bestcolor}\textbf{33.4} &
\cellcolor{bestcolor}\textbf{84.8} &
\cellcolor{bestcolor}\textbf{40.4} &
\cellcolor{bestcolor}\textbf{21.2} \\
\bottomrule \\ 
\label{tab:2}
\end{tabular}
\vspace{-1.9em}
\end{table*}

\section{Experiments}
\subsection{Experimental Settings}
\noindent \textbf{Implementation Details.} 
POMA-3D is pretrained for 20 epochs during the warm-up stage and 100 epochs in the main stage, with batch sizes of 1024 and 64, respectively. We adopt the AdamW~\cite{loshchilov2017decoupled} optimizer ($\beta_1$ = 0.9, $\beta_2$ = 0.98) with a learning rate of $1\times10^{-4}$, a warm-up of 500 steps, and cosine decay scheduling. The vision encoders are ViT-B/16 from FG-CLIP-Base~\cite{xie2025fg}. For the POMA-JEPA, the predictor depth is set to 2, and each point map view is assigned a single mask with a random scale in the range (0.15,\,0.2) and an aspect ratio in the range (0.75,\,1.5). All data construction, pretraining, and downstream fine-tuning are conducted on A100 (80\,GB) GPUs. More details are in the Appendix.

\definecolor{bestcolor}{RGB}{220,242,220}      % mint green – best
\definecolor{secondcolor}{RGB}{236,220,250}    % lavender – second best

% \vspace{-1em}
\begin{table*}[!ht]\addtolength{\tabcolsep}{-2.9pt}
\vspace{-1.0em}
\centering
\small

\caption{Scene retrieval results on ScanRefer~\cite{chen2020scanrefer}, Nr3D, and Sr3D~\cite{achlioptas2020referit3d}.
The metric R@M-N denotes recall@N for retrieving the correct 3D scene from M referring texts. 
All methods are evaluated in the zero-shot setting. 
\colorbox{bestcolor}{\textbf{Best}} 
and \colorbox{secondcolor}{second-best} results are highlighted.}
\vspace{-1em}
\begin{tabular}{lccccc|cccc|cccc}
\toprule
\multirow{2}{*}{Method} & 
\multirow{2}{*}{Modality} &
\multicolumn{4}{c|}{ScanRefer} &
\multicolumn{4}{c|}{Nr3D} &
\multicolumn{4}{c}{Sr3D} \\
\cmidrule(lr){3-6} \cmidrule(lr){7-10} \cmidrule(lr){11-14} 
& & 
R@1-1 & R@1-5 & R@5-1 & R@5-5 &
R@1-1 & R@1-5 & R@5-1 & R@5-5 &
R@1-1 & R@1-5 & R@5-1 & R@5-5 \\
\midrule
3D-ViSTA~\cite{zhu20233d} & C-PC &
0.48 & 2.27 & 0.24 & 2.03 &
0.45 & 0.60 & 0.15 & 0.60 &
0.33 & 1.15 & 0.33 & 1.48 \\

SceneVerse~\cite{jia2024sceneverse} & C-PC &
0.24 & 2.27 & 0.83 & 2.03 &
0.26 & 1.82 & 0.26 & 1.56 &
0.28 & 1.99 & 0.28 & 1.70 \\

FG-CLIP~\cite{xie2025fg} & RGB &
\cellcolor{secondcolor}5.10 & \cellcolor{secondcolor}16.4 & \cellcolor{secondcolor}14.9 & \cellcolor{secondcolor}42.2 &
\cellcolor{secondcolor}1.37 & \cellcolor{secondcolor}6.71 & \cellcolor{secondcolor}5.18 & \cellcolor{secondcolor}17.2 &
\cellcolor{secondcolor}1.35 & \cellcolor{secondcolor}6.42 & \cellcolor{secondcolor}1.86 & \cellcolor{secondcolor}10.1 \\

FG-CLIP~\cite{xie2025fg} & PM &
0.50 & 2.00 & 0.25 & 2.81 &
0.46 & 1.98 & 0.46 & 2.13 &
0.34 & 1.18 & 0.17 & 0.84 \\

\textbf{POMA-3D} & PM &
\cellcolor{bestcolor}\textbf{9.31} & \cellcolor{bestcolor}\textbf{27.9} & \cellcolor{bestcolor}\textbf{29.4} & \cellcolor{bestcolor}\textbf{59.4} &
\cellcolor{bestcolor}\textbf{8.10} & \cellcolor{bestcolor}\textbf{15.7} & \cellcolor{bestcolor}\textbf{15.0} & \cellcolor{bestcolor}\textbf{42.2} &
\cellcolor{bestcolor}\textbf{3.89} & \cellcolor{bestcolor}\textbf{14.0} & \cellcolor{bestcolor}\textbf{6.59} & \cellcolor{bestcolor}\textbf{20.7} \\
\bottomrule
\end{tabular}
\label{tab:3}
\vspace{-1.5em}
\end{table*}

\begin{figure*}[!ht]
    \centering
    \includegraphics[width=1.0\linewidth]{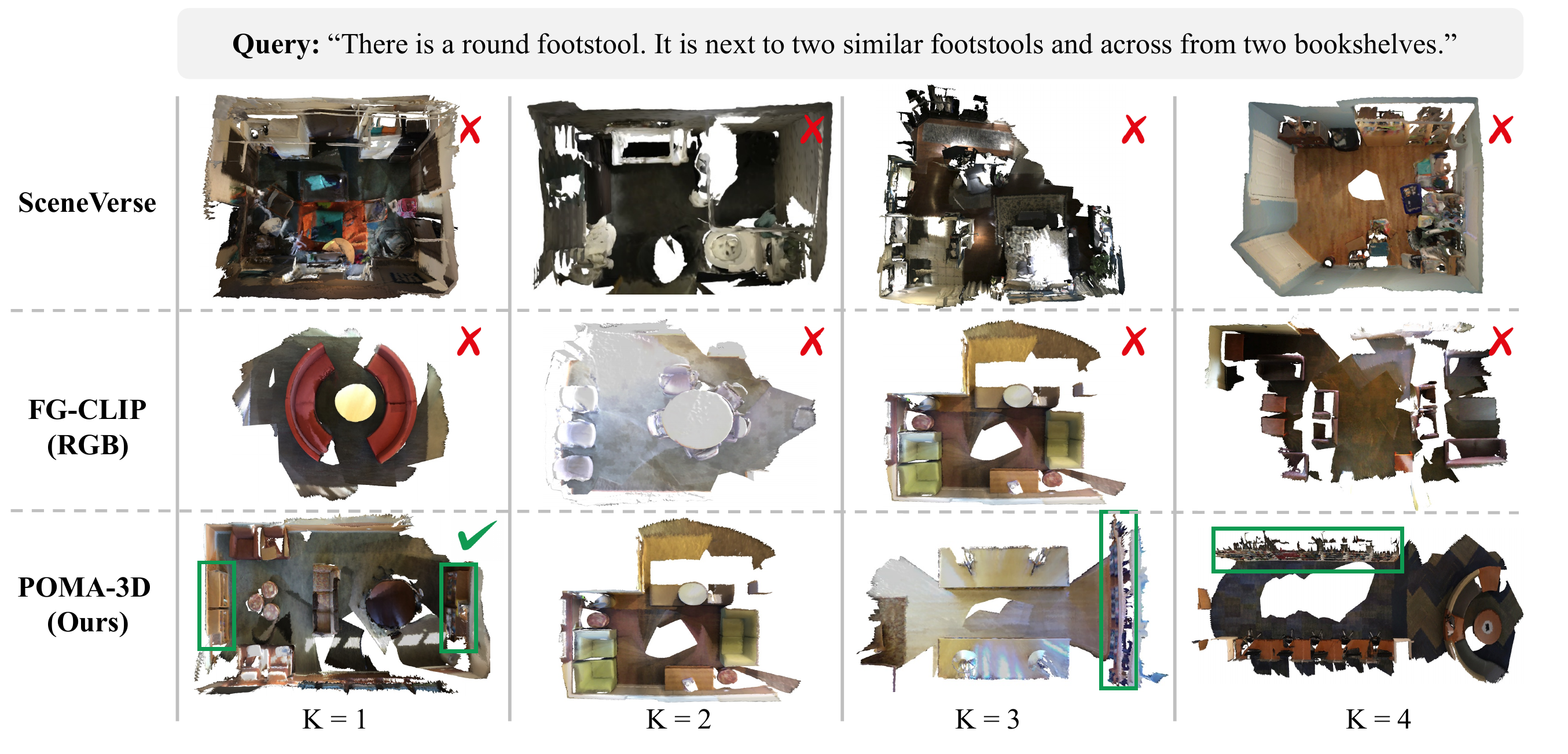}
    \vspace{-1.4em}
    \caption{\textbf{Qualitative scene retrieval results.} Top-4 candidates from each method are shown. For the given query, only POMA-3D retrieves the unique ground-truth scene, while others fail to return bookshelf-containing scenes. \textcolor{ForestGreen}{Green boxes} mark bookshelves.}
    \label{fig:3}
\end{figure*}

\begin{figure*}[!ht]
    \centering
    % \vspace{-0.5em}
    \includegraphics[width=1.0\linewidth]{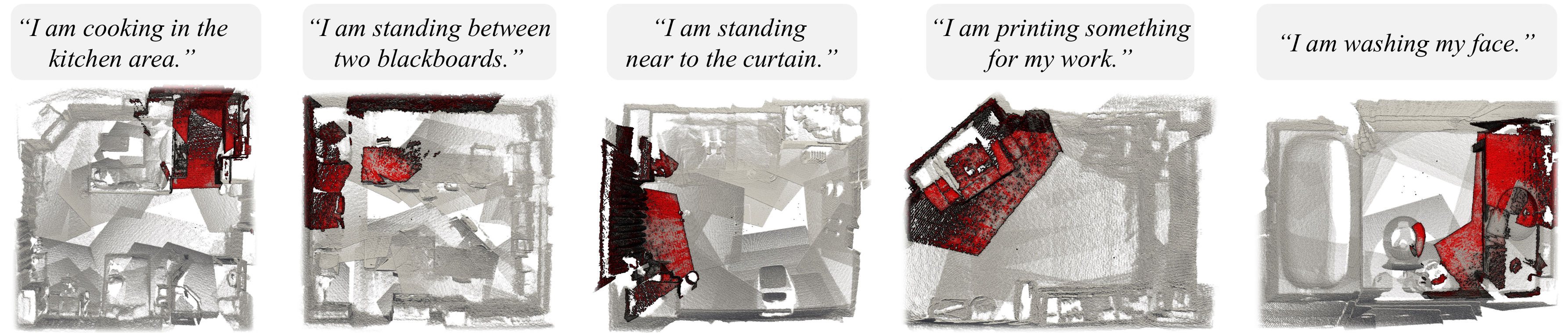}
    \vspace{-1.4em}
    \caption{\textbf{Qualitative coarse localization results.} 
    Top: text to describe the current agent's situation. Bottom: merged multi-view point maps, where \textcolor{red}{red} regions indicate the point map views retrieved by POMA-3D based on the text.}
    \label{fig:4}
    \vspace{-1.0em}
\end{figure*}

\noindent \textbf{3D QA Setting.} 
We evaluate our model on three benchmarks: the ScanQA~\cite{azuma2022scanqa} validation set, the SQA3D~\cite{ma2022sqa3d} test set, and the ScanNet split of Hypo3D~\cite{mao2025hypo3d}, which assess commonsense spatial, situated, and hypothetical reasoning, respectively. The Hypo3D dataset is divided into training, validation, and test sets with an 8:1:1 ratio based on scene IDs. Based on the POMA-3D encoder, we develop two baselines: a specialist model, POMA-3D$_{\text{spec}}$, and a generalist model, POMA-3D$_{\text{llm}}$. 
POMA-3D$_{\text{spec}}$ follows 3D-VisTA~\cite{zhu20233d} and SceneVerse~\cite{jia2024sceneverse}, consisting of a BERT language encoder and a QA head, and is fine-tuned with QA loss. POMA-3D$_{\text{llm}}$ aligns the POMA-3D encoder with the LLaVA-OV LLM via one epoch of LoRA fine-tuning, following the same protocol as SplatTalk~\cite{thai2025splattalk}. We compare POMA-3D$_{\text{spec}}$ and POMA-3D$_{\text{llm}}$ against three 3D LLMs, three 2D LLMs, and five specialist models. For fairness, all 2D LLMs, following POMA-3D, take 32-view images as input. 3D-VisTA~\cite{zhu20233d}, SceneVerse~\cite{jia2024sceneverse}, and SplatTalk~\cite{thai2025splattalk} are 3D VLL models fine-tuned on QA tasks and are most related to us. Since 3D-VisTA and SceneVerse additionally require object masks, evaluations on ScanQA and SQA3D employ masks generated by Mask3D~\cite{schult2022mask3d}, while Hypo3D uses ground-truth masks. For LLM-based models, we report exact match (EM@1) scores, and for specialist models, we report both EM@1 and EM@10 metrics.

\noindent \textbf{Embodied Navigation Setting.} 
The embodied navigation task is evaluated on the MSNN~\cite{linghu2024multi} dataset, which tests a model’s ability to infer the correct navigational direction given the 3D scene, agent’s situation, and task instruction (see Fig.~\ref{fig:1}). MSNN is divided into training, validation, and test sets with an 8:1:1 split. Models used for 3D QA are also applied to this task. Each sample provides answers at four- and eight-directional granularities (4-dir. and 8-dir.).

\noindent \textbf{Scene Retrieval Setting.} 
Ground-truth bounding boxes in existing 3D visual grounding (VG) datasets are defined on post-processed point clouds that are misaligned with point maps directly projected from depth maps, making direct evaluation of POMA-3D non-trivial. Instead, we repurpose existing 3D VG datasets, including ScanRefer~\cite{chen2020scanrefer}, Nr3D, and Sr3D~\cite{achlioptas2020referit3d}, for scene retrieval. In this task, the model retrieves the 3D scene given a scene description composed of referring texts from the dataset. Retrieval is performed by computing the similarity between the description embeddings and the mean-pooled point map embeddings, selecting the scene with the highest similarity. We compare POMA-3D against 3D-VisTA~\cite{zhu20233d}, SceneVerse~\cite{jia2024sceneverse}, and FG-CLIP~\cite{xie2025fg}. Performance is reported using R@M–N, where M is the number of referring texts composing the scene description and N is the Top-N recall.

\noindent \textbf{Coarse Localization Setting.} 
The view-level alignment enables POMA-3D to retrieve point map views matching the agent’s location (i.e., coarse localization). Specifically, the similarity between each point map view embedding within a scene and the situational text embedding (e.g., “I am sitting on the bed”) is computed. The Top-$K$ most similar point maps are retrieved. When all point maps in the scene are concatenated, the retrieved ones collectively highlight the agent’s active region in the world frame. We qualitatively evaluate coarse localization with $K=3$.

\subsection{Downstream Task Results}
\noindent \textbf{3D QA.} 
As shown in Tab.~\ref{tab:2}, our specialist model POMA-3D$_{\text{spec}}$ outperforms all evaluated specialist models on the SQA3D and Hypo3D datasets. However, it does not outperform all methods on ScanQA, likely because ScanQA contains a number of color-dependent questions. Comparison models leverage color inputs (e.g., colored point clouds or images), while our input is uncolored point maps that lack such cues. Additionally, POMA-3D$_{\text{spec}}$ exhibits consistent gains in EM@10, surpassing all models across datasets. 
In particular, it outperforms the state-of-the-art point cloud–based VLL model SceneVerse by 6.2\% on SQA3D and 4.5\% on Hypo3D. 
Even when compared to large 3D LLMs such as LEO and LLaVA-3D, POMA-3D$_{\text{spec}}$ still achieves noticeable improvements on Hypo3D. Relative to its FG-CLIP baseline, POMA-3D$_{\text{spec}}$ further improves EM@1 and EM@10 by around 2\% on all benchmarks, confirming the effectiveness of the proposed pretraining strategy. 
Furthermore, POMA-3D$_{\text{llm}}$ outperforms existing 2D LLMs on both SQA3D and Hypo3D. Compared to SplatTalk, POMA-3D$_{\text{llm}}$ achieves a 4\% improvement on SQA3D, illustrating the superiority of point map representation over 3DGS for situational reasoning.

\noindent \textbf{Embodied Navigation.}
Tab.~\ref{tab:2} demonstrates that among all models, POMA-3D$_{\text{spec}}$ achieves the highest performance on four-directional navigation, while POMA-3D$_{\text{llm}}$ performs best on eight-directional navigation. This indicates that POMA-3D’s advantage becomes more pronounced when requiring complex spatial reasoning beyond simple relationship and attribute recognition in traditional 3D QA.

\noindent \textbf{Scene Retrieval.} Tab.~\ref{tab:3} presents the quantitative results for scene retrieval. POMA-3D consistently outperforms its FG-CLIP baseline and all 3D VLL methods across datasets and metrics. Notably, prior approaches underperform the RGB-based FG-CLIP model, and in some metrics, even below its point map variant, despite FG-CLIP being pretrained solely on image data. This is likely because these methods align a newly trained point encoder with a BERT text encoder, failing to leverage the strong priors in pre-aligned CLIP. In contrast, POMA-3D builds upon the pretrained FG-CLIP, effectively extending its image–text alignment to point map–text alignment. Fig.~\ref{fig:3} further qualitatively demonstrates our method’s superiority. For a query requiring three footstools and bookshelves, only POMA-3D retrieves the single correct scene and returns bookshelf-containing results in three of the top four matches (green boxes), while all other methods fail to meet the query conditions.

\noindent \textbf{Coarse Localization.}
Qualitative results in Fig.~\ref{fig:4} show that POMA-3D accurately retrieves point map views describing the agent’s position from situational text containing multiple objects. For example, given “I am standing between two blackboards,” it correctly identifies the intersection area. It also performs well when no object references in queries, such as in “I am washing my face,” where it locates the basin area, demonstrating strong situational reasoning.

\subsection{Ablation Studies}
\noindent \textbf{Effect of Warmup with 2D Data.}
Tab.~\ref{tab:4} presents the ablations of POMA-3D components across 3D QA datasets. The first row, without any pretraining, corresponds to the FG-CLIP finetuned on the downstream tasks. Warmup pretraining on point maps derived from images improves EM@1 by 0.6\% on SQA3D and 1.3\% on Hypo3D. When combined with the main stage, the warmup further boosts ScanQA by 0.7\%. Hence, pretraining on 2D-derived point maps makes POMA-3D learn more robust 3D features.

\noindent \textbf{Effect of the Room-Level Scenes.}
Consistent with prior 3D VLL methods, incorporating room scenes benefits POMA-3D pretraining. As shown in the third row of Tab.\ref{tab:4}, room-level alignment with $\mathcal{L}_{\text{view}}$ and $\mathcal{L}_{\text{scene}}$ boots EM@1 across all benchmarks, surpassing all specialist models and several LLM-based methods on Hypo3D (See Tab.\ref{tab:2}).

\begin{table}[!t]\addtolength{\tabcolsep}{-2.5pt}
\centering
\small
\caption{
Ablation study of POMA-3D on 3D QA datasets.
}
\vspace{-1em}
\begin{tabular}{c|ccc|ccc}
\toprule
{Warmup} & 
\multicolumn{3}{c|}{{Main}} & 
\multicolumn{3}{c}{{EM@1}} \\
\cmidrule(lr){1-1} \cmidrule(lr){2-4} \cmidrule(lr){5-7}
$\mathcal{L}_{\text{view}}$ & 
$\mathcal{L}_{\text{view}}$ & $\mathcal{L}_{\text{scene}}$ & $\mathcal{L}_{\text{pjepa}}$ & 
{ScanQA } & {SQA3D } & {Hypo3D } \\
\midrule
\xmark & \xmark & \xmark & \xmark & 20.9 & 49.5 & 31.1 \\
\cmark & \xmark & \xmark & \xmark & 20.6 & 50.1 & 32.4 \\
\xmark & \cmark & \cmark & \xmark & 21.4 & 50.4 & 32.6 \\
\cmark & \cmark & \cmark & \xmark & 22.1 & 50.7 & 32.9 \\
\cmark & \cmark & \cmark & \cmark & \textbf{22.3} & \textbf{51.1} & \textbf{33.4} \\
\bottomrule
\end{tabular}
\label{tab:4}
% \vspace{-0.5em}
\end{table}

\noindent \textbf{Effect of POMA-JEPA.}
Tab.~\ref{tab:4} shows that incorporating the POMA-JEPA training objective consistently improves POMA-3D performance across all benchmarks, with the largest gains observed on SQA3D and Hypo3D. These results suggest that enforcing geometric consistency across multi-view features further strengthens POMA-3D’s capability in situated and hypothetical reasoning.

\begin{table}[t]\addtolength{\tabcolsep}{0.5pt}
\centering
\small
\caption{Ablation study of FG-CLIP and POMA-3D as visual encoders for LLaVA-OV. Both models were pretrained on SQA3D~\cite{ma2022sqa3d} and evaluated zero-shot on Hypo3D~\cite{mao2025hypo3d}.}
\vspace{-1em}
\begin{tabular}{lccc|c}
\toprule
{Visual Encoder} &  {Scale} & {Direction} & Semantic & {Overall} \\
\midrule
FG-CLIP~\cite{xie2025fg} & 44.1 & 15.2 & 33.3 & 30.3 \\
POMA-3D & \textbf{45.3} & \textbf{16.7} & \textbf{33.6}  & \textbf{30.9} \\
\bottomrule
\end{tabular}%
\label{tab:5}
\vspace{-1.5em}
\end{table}
\noindent \textbf{Zero-Shot Generalization of POMA-3D.} All results above evaluate POMA-3D under fine-tuning on downstream benchmarks. Tab.~\ref{tab:5} further presents cross-dataset results, comparing FG-CLIP and POMA-3D as visual encoders within LLaVA-OV. Both models are fine-tuned on SQA3D and evaluated zero-shot on Hypo3D. The POMA-3D–based LLaVA-OV achieves higher overall EM@1, with gains of 1.2\% and 1.5\% on scale and direction questions, respectively (i.e., spatial questions). These results demonstrate that POMA-3D features transfer effectively across tasks.

\begin{figure}
    \centering
    \includegraphics[width=1.0\linewidth]{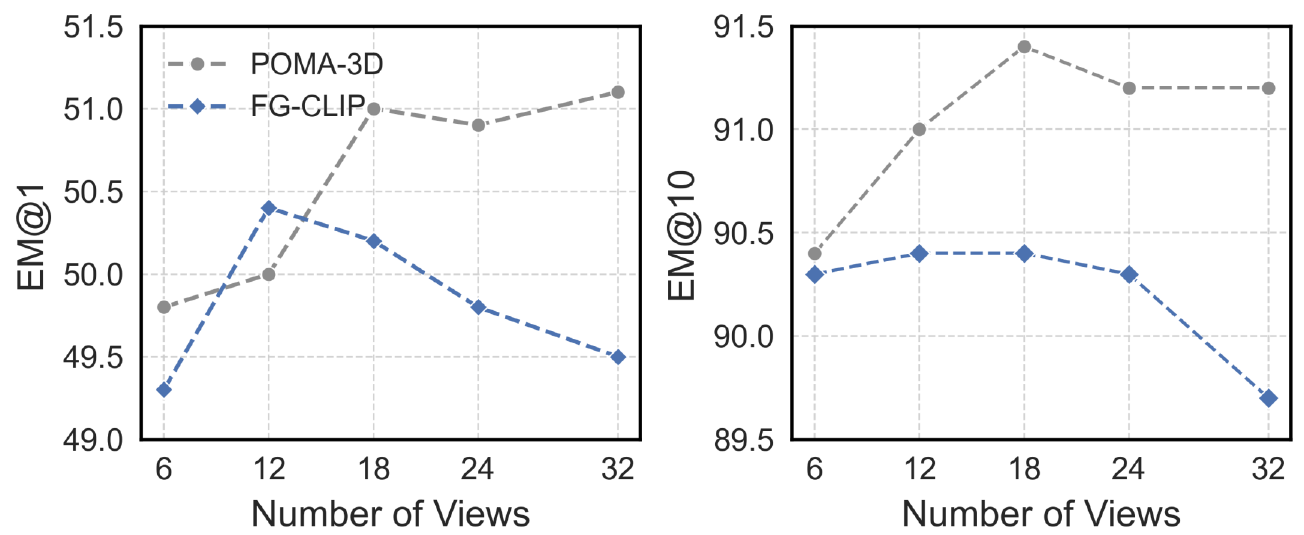}
        \vspace{-1.5em}
    \caption{Comparison of POMA-3D and its baseline FG-CLIP with varying numbers of views on SQA3D~\cite{ma2022sqa3d}.}
    \label{fig:5}
\end{figure}

\begin{table}[t]\addtolength{\tabcolsep}{0.5pt}
\centering
\small
% \vspace{-0.5em}
\caption{3D QA performance of pretrained and aligned FG-CLIP models using depth maps and point maps as input.}
\vspace{-1em}
\begin{tabular}{lcccc}
\toprule
Modality & {$\mathcal{L}_{\text{view}}$ $\mathcal{L}_{\text{scene}}$} & {ScanQA} & {SQA3D} & {Hypo3D} \\
\midrule
% Image & \xmark & 21.4 & 51.0 & 32.0 \\
% \midrule
\multirow{2}{*}{Depth Map} 
& \xmark & 20.1 & 49.1 & 31.0 \\
& \cmark & 21.0 & 50.1 & 31.5 \\
\midrule
\multirow{2}{*}{Point Map} 
& \xmark & 20.9 & 49.5 & 31.1 \\
& \cmark & \textbf{21.4} & \textbf{50.4} & \textbf{32.6} \\
\bottomrule
\end{tabular}
\label{tab:6}
 \vspace{-1.0em}
\end{table}
\noindent \textbf{Effect of the Number of Point Map Views.} 
We evaluate POMA-3D under different numbers of views used during finetuning on SQA3D task. As shown in Fig.~\ref{fig:5}, POMA-3D consistently outperforms the FG-CLIP across all view counts on EM@1 and EM@10. Unlike FG-CLIP, which plateaus or declines with more views, POMA-3D improves steadily, showing that each point map view in POMA-3D provides complementary features.

\noindent \textbf{Comparing Point Map with Other 3D Modalities.}
This experiment hypothesizes that the point map is a more effective 3D modality for representation learning. Tab.~\ref{tab:6} compares pretrained FG-CLIP models using point maps and depth maps as inputs for 3D QA. Point maps yield consistently better results, suggesting that 2D priors transfer more effectively to point maps. This advantage persists even when both modalities are aligned with $\mathcal{L}_{\text{view}}$ and $\mathcal{L}_{\text{scene}}$. Moreover, point map features receive additional gains from POMA-JEPA, while depth maps do not. Tab.~\ref{tab:2} further shows that POMA-3D surpasses point cloud– and GS-based methods. Altogether, these findings highlight point maps as a more promising 3D modality for feature learning.

% Table

% Full-width simplified table (without Scan2Cap columns)

% \begin{table}[t]
% \centering
% \caption{Ablation study on the effect of using cls and patch token. We report accuracy (\%) on three datasets: ScanQA, SQA3D, and Hypo3D.}
% \label{tab:ablation_views}
% \resizebox{0.9\linewidth}{!}{%
% \begin{tabular}{ccccc}
% \toprule
% \textbf{Token Type} & \textbf{Backbone} & \textbf{ScanQA} & \textbf{SQA3D} & \textbf{Hypo3D} \\
% \midrule
% cls & FG-CLIP + LLaVA-OV & 00.0 & 00.0 & 00.0 \\
% cls  & POMA-3D &00.0 & 00.0 & 00.0 \\
% Patch  & LVLM & 00.0 & 00.0 & 00.0 \\
% Patch  & LVLM &00.0 & 00.0 & 00.0 \\
% Patch  & LVLM &00.0 & 00.0 & 00.0 \\
% \bottomrule
% \end{tabular}%
% }
% \end{table}

% FT 52. 

% Baseline 52 

% r 16 alpha 64 lr 1e-5
% r 256 alph 512 lr 1e-4
% PM  scanqa 

% Hypo3D 27.41
% SQA3D 40.52

% scanqa pretrain 512 1024
% /gpfs/home/ym621/UniPointMap/lmms-finetune/checkpoints/sqa_llava-onevision-7b-ov_poma_r256a512_lora-True_qlora-False

% 30.89
% 30.34

% Scene Caption -> Scene

% Scene Caption -> multi-view point map POMA-3D FG-CLIP 

% coarse localization 

%% file: sec/6_conclusion.tex
\section{Conclusion}
In this work, we introduce POMA-3D, the first point map-based self-supervised 3D model. Pretrained on the ScenePoint dataset using view-to-scene vision–language alignment and POMA-JEPA objectives, POMA-3D learns robust point map representations that generalize to diverse 3D tasks. The learned features strengthen both lightweight specialist and large generalist 3D models. Importantly, POMA-3D effectively inherits priors from 2D foundation models and benefits from large-scale image datasets. We believe this work paves the way to scalable 3D understanding. 

\noindent\textbf{Limitation.} Since point maps contain only coordinates, future work should explore fusing them with color features to build a unified scene representation.